\documentclass[lettersize,journal]{IEEEtran}

\usepackage{amsmath,amsfonts,amssymb}
\usepackage{booktabs}
\usepackage[inline]{enumitem}
\usepackage{textcomp}
\usepackage{stfloats}
\usepackage{url}
\usepackage{verbatim}
\usepackage{graphicx}
\usepackage{multirow}
\usepackage{subfigure}
\usepackage{float}
\usepackage{comment}
\usepackage{bbm}
\newcommand{\etal}{\textit{et al.}}
\usepackage{tablefootnote}
\usepackage{multicol,lipsum}

\usepackage[backend=bibtex,style=numeric,natbib=true,citestyle=numeric,sorting=none,backref=false,backrefstyle=three,maxcitenames=2]{biblatex}
\bibliography{refs}
\usepackage[breaklinks=true,colorlinks,bookmarks=false]{hyperref}
\AtBeginBibliography{\small}
\defbibfilter{appendixOnlyFilter}{
  segment=1 
  and not segment=0 
}
\usepackage[capitalise,noabbrev]{cleveref}
\usepackage[pagewise]{lineno}
\usepackage[normalem]{ulem}

\usepackage{arydshln}

\usepackage{color}
\usepackage[
]{xcolor}

\usepackage{fancyhdr}
\fancypagestyle{copyright}{%
    \fancyhf{}
    \cfoot{\footnotesize%
    }
}

\begin{document}
\title{Feature Guided Masked Autoencoder for Self-supervised Learning in Remote Sensing
}

\author{Yi Wang,~\IEEEmembership{Student Member,~IEEE}, Hugo Hernández Hernández, Conrad M Albrecht,~\IEEEmembership{Member,~IEEE}, \\ Xiao Xiang Zhu,~\IEEEmembership{Fellow,~IEEE}

\thanks{Y. Wang, H. H. Hernández and X. Zhu are with the chair of Data Science in Earth Observation, Technical University of Munich (TUM), Germany.

Y. Wang and C. M. Albrecht are with Remote Sensing Technology Institute, German Aerospace Center (DLR), Germany.

X. Zhu is with Munich Center for Machine Learning, Munich, Germany.

Codes and pretrained models are available at \url{https://github.com/zhu-xlab/FGMAE}. The collected EuroSAT-SAR dataset is available at \url{https://huggingface.co/datasets/wangyi111/EuroSAT-SAR}.
}
}



\maketitle
\begin{abstract}
Self-supervised learning guided by masked image modelling, such as Masked AutoEncoder (MAE), has attracted wide attention for pretraining vision transformers in remote sensing. However, MAE tends to excessively focus on pixel details, thereby limiting the model's capacity for semantic understanding, in particular for noisy SAR images. In this paper, we explore spectral and spatial remote sensing image features as improved MAE-reconstruction targets. We first conduct a study on reconstructing various image features, all performing comparably well or better than raw pixels. Based on such observations, we propose \textit{Feature Guided Masked Autoencoder} (FG-MAE): reconstructing a combination of Histograms of Oriented Graidents (HOG) and Normalized Difference Indices (NDI) for multispectral images, and reconstructing HOG for SAR images. Experimental results on three downstream tasks illustrate the effectiveness of FG-MAE with a particular boost for SAR imagery. Furthermore, we demonstrate the well-inherited scalability of FG-MAE and release a first series of pretrained vision transformers for medium resolution SAR and multispectral images.
\end{abstract}


\thispagestyle{copyright}

\vspace{-1em}
\section{Introduction}

\IEEEPARstart{S}{elf-supervised} Learning has brought breakthroughs to the remote sensing (RS) community with the ability to learn generic representations from large-scale unlabeled data \cite{wang2022self}. The pretrained encoders (recently also called foundation models) can then be transferred to various downstream applications. While convolutional neural networks have been long studied as model backbones with contrastive learning \cite{jean2019tile2vec}, there is a growing trend of pretraining vision transformers (ViT) \cite{dosovitskiy2020image} with masked image modeling (MIM), particularly, masked autoencoder (MAE) \cite{he2022masked} and its variants \cite{cong2022satmae}.

\begin{figure}[]
  \centering
  \includegraphics[width=1.0\linewidth]{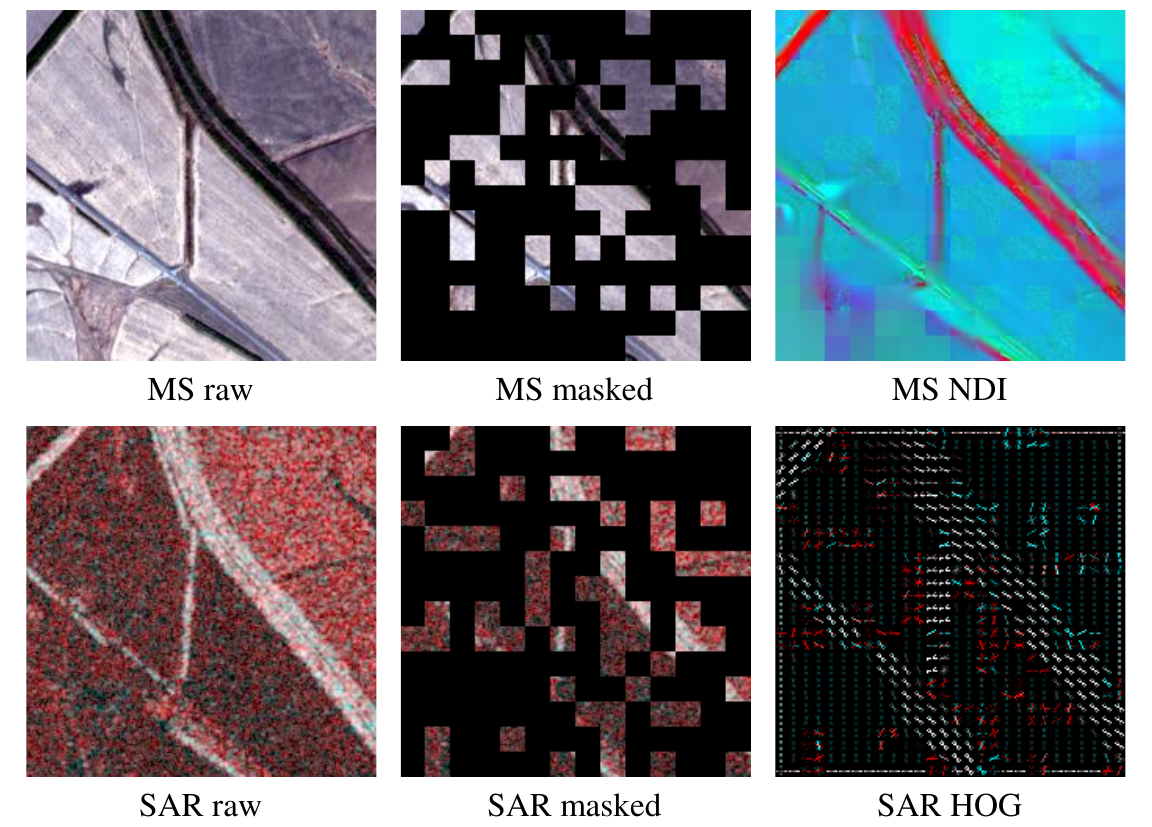}
  \caption{Sample data of the proposed FG-MAE method---columns from left to right: Sentinel-2 multispectral (MS) and Sentinel-1 (SAR) imagery, masked model inputs, model-reconstructed features (HOG: Histogram of Gradients, NDI: Normalized Difference Index). False color of the raw SAR image is coded by [VV, VH, (VV+VH)/2]. False color of the reconstructed MS NDI is coded by [NDVI, NDWI, NDBI].}
  \vspace{-1em}
  \label{fig:features}
\end{figure}

MAE works as masking some patches of an input image, encoding the unmasked patches, and reconstructing the masked patches. Such asymmetric encoder-decoder design makes it highly efficient compared to contrastive learning. However, reconstructing raw input makes MAE over-focus pixel details, sensible to artifacts and noise, and potentially diverting attention from high-level semantic representations. These challenges are exacerbated in synthetic aperture radar (SAR) scenarios, in which the exisitence of speckle noise, which appears as a granular disturbance and usually modeled as a multiplicative noise, limits MAE's performance.

In this work, we propose a new simple variant of MAE for RS imagery, termed Feature Guided Masked Autoencoder (FG-MAE), by replacing raw images with image features as reconstruction targets. Looking back at traditional RS image analysis, human designed feature descriptors (e.g. edge or vegetation index) have been widely used to extract semantic information of the Earth's surface \cite{ali2001using,lunetta2006land}. These image features incorporate expert knowledge, and can guide the model's learning process when introduced to MAE. To demonstrate that, we conduct a study on popular features for multispectral and SAR imagery: 1) CannyEdge \cite{canny1986computational}, 2) histograms of oriented gradients (HOG) \cite{dalal2005histograms}, 3) scale-invariant feature transform (SIFT) \cite{lowe2004distinctive}, and 4) normalized difference indices (NDI) \cite{pettorelli2013normalized,gao1996ndwi,zha2003use}. We show that each of these features alone works comparably well or even better than the original MAE.

We then search for the best candidates among the popular features, and propose FGMAE-MS and FGMAE-SAR. For multispectral imagery, we combine the spatial feature HOG and the spectral feature NDI, using two separate prediction heads at the end of the decoder. This combination allows the spatial and spectral features to complement each other. For SAR imagery, we simply use HOG to enhance spatial information and reduce the influence of speckle noise. We evaluate FG-MAE on scene classification and semantic segmentation downstream tasks with BigEarthNet-MM \cite{sumbul2021bigearthnet}, EuroSAT \cite{helber2019eurosat} and DFC2020 \cite{schmitt2020ieee} datasets for both multispectral and SAR images. For EuroSAT, we match the geocoordinates of EuroSAT-MS and collect the EuroSAT-SAR dataset. Results demonstrate the effectiveness of FG-MAE on all tasks, particularly in SAR scenarios. In addition, FG-MAE remains as efficient as MAE, making it possible to scale up to big foundation models. We show that both FGMAE-MS and FGMAE-SAR scale well up to ViT-Huge with 0.7B parameters under linear evaluation protocols.

Our main contributions are listed as follows:
\begin{itemize}
\item We demonstrate the effectiveness of using RS image features as reconstruction targets for masked image modeling based self-supervised learning;
\item We propose FG-MAE, a new variant of MAE that works well for both multispectral and SAR imagery;
\item We show the benefits of FG-MAE pretrained models on three popular MS\&SAR datasets;
\item We verify the scalability of FG-MAE, and release a first series of pretrained ViTs for multispectral and SAR images with parameter sizes ranging from 22M to 0.7B.
\end{itemize}

\vspace{-0.3em}
\section{Related work}
\noindent{\textbf{Masked image modeling for self-supervised learning}} \hspace{0.3em}
Masked image modeling (MIM) is a recent family of generative self-supervised learning that focus on pretraining vision transformers by reconstructing the masked input, such as iGPT \cite{chen2020generative}, BEiT \cite{bao2021beit} and SimMIM \cite{xie2022simmim}. Of particular interest, MAE \cite{he2022masked} drew wide attention with substantial improvements on fine-tuning downstream tasks and efficient pretraining. 

Our work, FG-MAE, is a simple variant of MAE. Instead of reconstructing raw images, we propose to reconstruct features that are better suited for RS imagery. FG-MAE is also closely related to MaskFeat \cite{wei2022masked}, where the authors introduced masked feature prediction for self-supervised video representation learning. We propose to use the asymmetric encoder-decoder structure of MAE for efficiency, and explore the best features for multispectral and SAR imagery.

\vspace{0.5em}
\noindent{\textbf{Masked image modeling in remote sensing}} \hspace{0.3em}
Most existing MIM works in RS are based on MAE \cite{fuller2022satvit,sun2022ringmo}. SatViT \cite{fuller2022satvit} presents the benefits of a straightforward implementation of MAE on satellite images. Wang~\etal~\cite{wang2022land} showcased the potential of MAE on PolSAR images. RingMo \cite{sun2022ringmo} modified the masking strategy by reversing some pixels in the masked patches to avoid complete lost of small objects. MAEST \cite{ibanez2022masked} implemented MAE on hyperspectral images with spectral masking. SatMAE \cite{cong2022satmae} proposed temporal and spectral masking and positional encoding in multispectral remote sensing time series. Scale-MAE \cite{reed2023scale} introduced ground sampling distance positional encoding and multiscale reconstruction to capture the geospatial scale information of RS images. Our work differs from all aforementioned approaches by improving MAE for RS imagery from the perspective of reconstructing image features as targets.

\vspace{0.5em}
\noindent{\textbf{Exploiting image features in remote sensing}} \hspace{0.3em}
Image feature descriptors play a big role in traditional RS image analysis. The normalized difference indices have long been used for Earth surface monitoring since the last century \cite{carlson1997relation,gao1996ndwi}. Similarly, spatial features like HOG are widely used as input to machine learning algorithms \cite{torrione2013histograms}. In this work, we revisit these well-known human-designed features and let them be learned by deep neural networks. This approach leverages the expertise of human analysts to guide the training process and facilitate the learning of better representations.

\vspace{-0.3em}
\section{Methodology}
\label{sec:methodology}

\begin{figure*}[ht]
  \centering
  \includegraphics[width=1.0\linewidth]{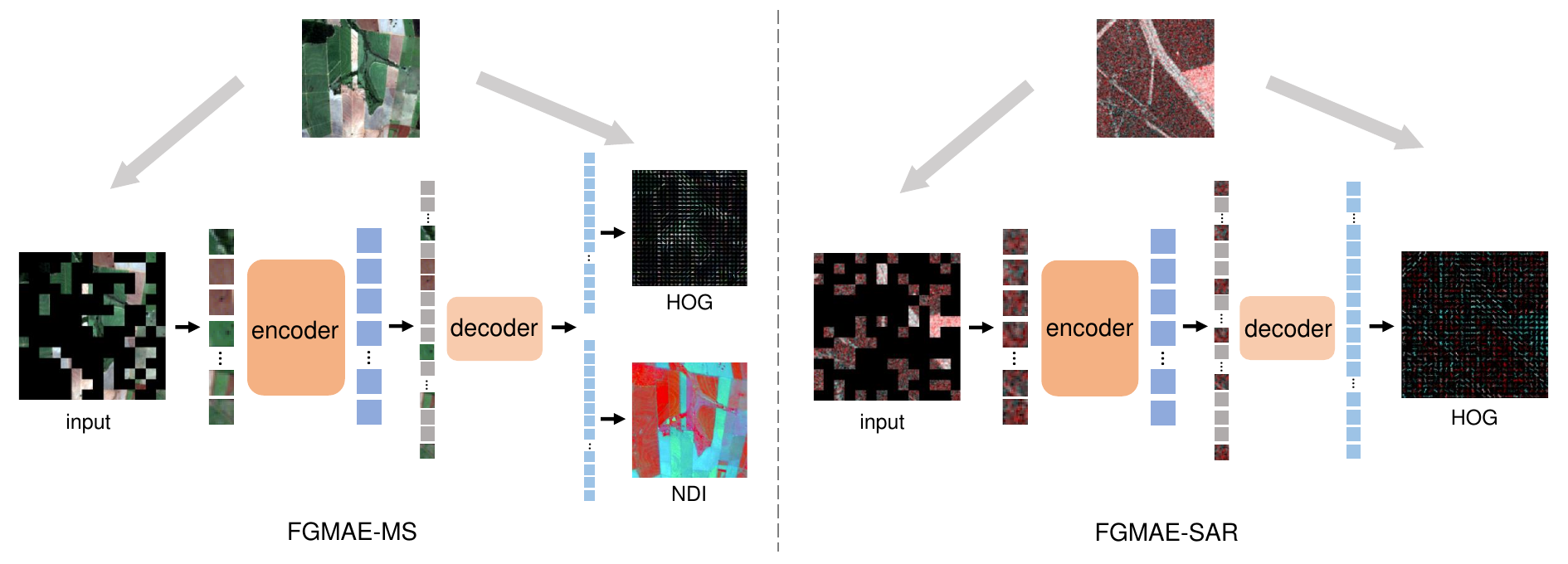}
  \caption{The general structure of the proposed FG-MAE method. We replace the reconstruction target of MAE \cite{he2022masked} by remote sensing image features.}
  \label{fig:fgmae}
\vspace{-1em}
\end{figure*}

Our proposed FG-MAE is a simple variant of MAE \cite{he2022masked} that replaces the reconstruction target with RS image features. As is illustrated in Figure \ref{fig:fgmae}, the image is divided into non-overlapping patches, and a random subset of these patches are masked out. The remaining visible patches are sent through the ViT encoder. The full set of encoded visible patches and learnable mask tokens are fed into the lightweight ViT decoder to reconstruct target features. During training, mean squared error or L2 loss is minimized only on masked patches. In the following subsections, we will discuss different feature candidates in \ref{subsec:feature}, and present the specific target designs for multispectral and SAR imagery in \ref{subsec:fgmae}, respectively.

\vspace{-0.3em}
\subsection{Target features}
\label{subsec:feature}
We consider two categories and four types of RS image features: spatially, 1) CannyEdge \cite{canny1986computational}, 2) HOG \cite{dalal2005histograms}, and 3) SIFT \cite{lowe2004distinctive}; spectrally, 4) NDI, including vegetation index \cite{pettorelli2013normalized}, water index \cite{gao1996ndwi} and built-up index \cite{zha2003use}.

\vspace{0.5em}
\noindent{\textbf{CannyEdge}}
\hspace{0.3em}
CannyEdge \cite{canny1986computational} is an edge detection algorithm that identifies the edges in an image by tracing the gradient of pixel intensities. The algorithm works by convolving the image with a Gaussian filter to reduce noise, and then computing the gradient magnitude and direction of each pixel. Non-maximum suppression is applied to suppress non-max edge contributors, and edges are detected by applying a Hysteresis threshold to the gradient magnitude.

Edge descriptors can simplify complex images by highlighting object boundaries, facilitating object identification and tracking in computer vision algorithms. As one of the most popular algorithms in this family, CannyEdge has the ability to accurately detect edges while minimizing false positives. It can also adapt to changes in lighting and contrast, which can often cause issues for other edge detection algorithms. Additionally, CannyEdge is able to accurately detect edges regardless of their orientation or position within the image. This makes it a powerful tool for remote sensing applications \cite{liu2004automated}.

CannyEdge is easy to compute in any deep learning framework by convolution, non-max suppression and thresholding. We use the filter toolbox of kornia \cite{riba2020kornia} to extract the edges as MAE targets (one edge map from one image channel). The same process as reconstructing the raw image follows, including patchifying and normalization within each small patch.

\vspace{0.5em}
\noindent{\textbf{HOG}}
\hspace{0.3em}
Histograms of Oriented Gradients \cite{dalal2005histograms} is a feature descriptor to describe the distribution of gradient orientations within a local subregion of an image. The algorithm calculates the magnitudes and orientations of gradients at each pixel using gradient filtering. Then, the gradients within each small local window are accumulated into normalized orientation histogram vectors voted by gradient magnitudes.

HOG is able to capture local shapes and appearances while being partially invariant to geometric changes. HOG is also invariant to photometric changes, as image gradients and local contrast normalization absorb brightness and foreground-background contrast variation. Unlike CannyEdge, HOG does not focus solely on edges but provides information about the magnitude and orientation of edge gradients.

HOG can be implemented similarly to CannyEdge as a two-channel convolution to generate gradients, followed by histogramming and normalization. We follow the implementation of MaskFeat \cite{wei2022masked} that writes HOG as (weight-fixed) neural network modules. Each channel of the raw image provides one HOG feature. The histograms of masked patches are then flattened and concatenated into a 1-D vector as the target feature.

\vspace{0.5em}
\noindent{\textbf{SIFT}}
\hspace{0.3em}
Scale-invariant feature transform (SIFT) \cite{lowe2004distinctive} is a feature descriptor that is used to extract distinctive and invariant local features from images. It works by detecting key points in an image that are invariant to scale, rotation, and illumination changes. Once the key points are detected, SIFT computes a descriptor for each key point by extracting the local image gradient orientations and magnitudes. These gradients are then transformed into a histogram of orientations, which is used to create a feature vector that describes the local image patch around the key point. 

The SIFT descriptor is robust against scale, rotation, illumination, and noise, making it applicable for a wide range of applications like image registration \cite{ma2016remote}. However, the complicated workflow of key point detectors and feature descriptors make it difficult for the model to learn. Another specific issue is that instead of region-based features, SIFT provides point-based features that do not align well with a standard ViT model design. Accordingly, it is tricky to integrate the famous SAR-SIFT \cite{dellinger2014sar} algorithm for SAR images. How to efficiently deal with the dynamic key points and the model's learning capacity remains a challenging task for future research. As a preliminary showcase in this work, we simplify the key point detection process by computing SIFT descriptor densely over the image. We utilize the feature toolbox of kornia \cite{riba2020kornia} to calculate dense SIFT features. Due to memory constraints, we perform the calculation using grayscale images.

\vspace{0.5em}
\noindent{\textbf{NDI}}
\hspace{0.3em}
Normalized Difference Indices (NDI) is a technique used to identify one type of ground objects by quantifying the differences between two spectral bands. It is often used in remote sensing applications such as changes in vegetation health or soil moisture levels. NDI works by calculating the ratio of the difference between two feature-sensitive spectral bands to their sum. This ratio is then normalized to a range between -1 and 1, where values closer to 1 indicate an increase in the feature of interest. 

NDI is a simple and effective way to detect changes in vegetation health or soil moisture levels, as it is sensitive to changes in the reflectance of different spectral bands. Three most popular NDIs are normalized difference vegetation index (NDVI), normalized difference water index (NDWI), and normalized built-up index (NDBI):

\begin{equation}
    NDVI = \frac{NIR-R}{NIR+R}
\end{equation}
\begin{equation}
    NDWI = \frac{G-NIR}{G+NIR}
\end{equation}
\begin{equation}
    NDBI = \frac{SWIR-NIR}{SWIR+NIR}
\end{equation}

\noindent where $NIR$ represents near infrared, $R$ represents red, $G$ represents green, and $SWIR$ represents short wave infrared. In this work, we calcuate the three indices for each pixel and concatenate them into a three-channel target image.

\vspace{0.5em}
In our experiments, we demonstrate that all above features serve as good reconstruction targets to replace raw images. Results will be discussed in \ref{subsec:feature-ablation}, where we perform a study on separately reconstructing the above features and evaluate corresponding downstream performances.

\subsection{FGMAE-MS / SAR}
\label{subsec:fgmae}

We then develop our proposed self-supervised methods, FG-MAE, based on the feature study. We consider two popular modalities in RS, multispectral imagery and polarimetric SAR imagery. For multispectral imagery, we combine spatial feature HOG and spectral feature NDI to complement each other; for SAR, we select HOG for its computational efficiency and noise robustness. 

As is shown in Figure \ref{fig:fgmae}, we retain the asymmetric encoder-decoder structure of MAE while modifying the reconstruction targets. Specifically, for FGMAE-SAR, the augmented raw images with shape (B,2,W,H) are divided into L non-overlapping patches with shape (B,L,w,h), of which $L_{m}$ random patches are masked out. The remaining visible patches with shape (B,$L-L_{m}$,w,h) are flattened to (B,$L-L_{m}$,w*h), processed with a linear embedding layer to (B,$L-L_{m}$,$K_{en}$) and passed through the ViT encoder. The encoded visible patches have shape (B,$L-L_{m}$,$K_{en}$). At the beginning of the decoding process, a linear layer is used to embed encoded patches to (B,$L-L_{m}$,$K_{de}$). They are then combined with mask tokens to (B,L,$K_{de}$) as input to a lightweight ViT decoder. The last layer of the decoder is a linear layer that converts the decoded patches to HOG predictions with shape (B,L,$K_{out}$), where $K_{out}$ is defined by HOG window size, number of bins and input channel numbers. 

While mostly similar for FGMAE-MS, the last layer of the decoder is replaced by two parallel linear layers, one outputting HOG and the other NDI. Note that for both modalities the outputs cover all patches, and only the masked ones are counted in the L2 loss calculation.

\section{Experimental setup}

\subsection{Self-supervised pretraining}

\noindent{\textbf{Dataset}}
\hspace{0.3em}
We pretrain vision transformers on Sentinel-1 GRD and Sentinel-2 L1C products of SSL4EO-S12 dataset \cite{wang2023ssl4eo}. The dataset is sampled from 250K locations around the world. Each location has four images from four seasons with size 264$\times$264 and ground sampling distance 10m. The multispectral images have 13 channels, and the SAR images have 2 channels.

\vspace{0.5em}
\noindent{\textbf{Data augmentation}}
\hspace{0.3em}
One image from a random season is selected for one location, followed by RandomResizedCrop to 224$\times$224 and RandomHorizontalFLip as the data augmentations.

\vspace{0.5em}
\noindent{\textbf{Model architecture}}
\hspace{0.3em}
We adopt the architecture design of MAE \cite{he2022masked}, which includes a regular ViT encoder (by default ViT-S/16 unless specifically noted) and a lightweight ViT decoder. Only the encoder is transferred to downstream tasks. The masking ratio is set to 70\% as recommended in \cite{wang2023ssl4eo}.

\vspace{0.5em}
\noindent{\textbf{Optimization}}
\hspace{0.3em}
We pretrain ViTs with batchsize 256 for 100 epochs. We use the AdamW optimizer \cite{loshchilov2018decoupled} with weight decay 0.05 and a basic learning rate 1.5e-4. The learning rate is warmed up for 10 epochs, and then decayed with a cosine schedule. Training is distributed across four NVIDIA A100 GPUs and takes about 7 hours for multispectral and 4 hours for SAR.

\subsection{Transfer learning} 

\noindent{\textbf{Dataset}}
\hspace{0.3em}
The pretrained models are transferred to scene classification and semantic segmentation downstream tasks for both multispectral and SAR imagery. For
\begin{itemize}
    \item \textit{scene classification}, we evaluate EuroSAT \citep{helber2019eurosat} (single-label land cover classification) and BigEarthNet-MM \citep{sumbul2021bigearthnet} (multi-label land cover classification) via linear probing (freeze encoder) and end-to-end fine tuning.
    \item \textit{semantic segmentation}, we evaluate DFC2020 \citep{schmitt2020ieee} (land cover segmentation) via fine tuning.
\end{itemize}

BigEarthNet-MM and DFC2020 have both multispectral and SAR images available. For BigEarthNet-MM, we use the 19-class labels, and follow the official train/val/test splits. For DFC2020, we use the 10-class high-resolution segmentation labels, and adjust the official test/validation data for 5128 training and 986 testing images. For EuroSAT, we perform a random 80\%/20\% train/test split.

Since EuroSAT has only RGB and multispectral images, we collected EuroSAT-SAR by pairing the published EuroSAT-MS from Sentinel-1 GRD products. This is done by matching the geocoordinates of EuroSAT-MS images and downloading the corresponding patches with Google Earth Engine \cite{gorelick2017google}. Because EuroSAT-MS has no exact collection time information, we performed a rough year-level match based on the publication time. In the end, we performed a manual check on random patches for the semantic correctness.

\vspace{0.5em}
\noindent{\textbf{Data augmentation}}
\hspace{0.3em}
We follow a common practice to use RandomResizedCrop (scale 0.2 to 1.0, resized to 224$\times$224) and RandomHorizontalFlip as data augmentations for all linear probing experiments. For BigEarthNet-MM, we set the smallest crop scale as 0.8 to avoid cutting out too many objects for the multilabel task. For DFC2020, we set the resized image size 256$\times$256 following MAE \cite{he2022masked}. For fine tuning experiments, we add mixup \cite{zhang2018mixup} augmentation. The multispectral images of BigEarthNet-MM (Sentinel-2 L2A) are zero-padded to 13 channels to match the pretrained models.

\vspace{0.5em}
\noindent{\textbf{Model architecture}}
\hspace{0.3em}
We use standard ViTs for scene classification on BigEarthNet-MM and EuroSAT. For semantic segmentation on DFC2020, we use UperNet \cite{xiao2018unified} with ViT backbones following MAE \cite{he2022masked}. 

\vspace{0.5em}
\noindent{\textbf{Optimization}}
\hspace{0.3em}
For BigEarthNet-MM, we minimize MultiLabelSoftMargin loss. The batchsize is set to 256. For linear probing, we train SGD optimizer without weight decay for 50 epochs. For fine tuning, we train AdamW optimzier with weight and layer decay for 20 epochs. The learning rate is 0.5 with cosine decay for linear probing, and 1e-3 with cosine decay and 3-epoch warm-up for fine tuning.

For EuroSAT, we minize cross entropy loss. The batchsize is set to 256. For linear probing, we train SGD optimizer with weight decay 0.001 for 50 epochs. For fine tuning, we train AdamW optimzier with weight and layer decay for 20 epochs. The learning rate is 0.1 with cosine decay for linear probing, and 1e-3 with cosine decay and 3-epoch warm-up for fine tuning.

For DFC2020, we use the RSI-Segmentation library \cite{xiong2022earthnets} for fine tuning. We minimize cross entropy loss for 40k iterations with batchsize 8. We use AdamW optimizer with layer decay. The basic learning rate is 1e-4, which is warmed up for 500 iterations and then polynomial-decayed. 

\vspace{0.5em}
\noindent{\textbf{Evaluation metrics}}
\hspace{0.3em}
We use mean average precision (mAP) and F1 score for the evaluation of BigEarthNet-MM. Overall accuracy (OA) and class-wise average accuracy (AA) are used for EuroSAT. For DFC2020, we evaluate overage accuracy (OA), average accuracy (AA) and mean intersection over union (mIoU).

\section{Results}
\label{sec:results}

\subsection{FG-MAE: target features}
\label{subsec:feature-ablation}
We first conduct a study on replacing raw image with different target features in MAE for both multispectral and SAR imagery. We pretrain ViTs on SSL4EO-S12 and transfer them to a 10\% subset of BigEarthNet-MM. As shown in Table \ref{tab:feature-ms}, all features perform comparably well to the raw image (MAE) under both linear probing and fine tuning settings in multispectral imagery. HOG is even better than the raw image for both settings. This proves the effectiveness of reconstructing image features as a new variant of MAE.

\begin{figure*}
    \centering
    
    \includegraphics[width=1.0\linewidth]{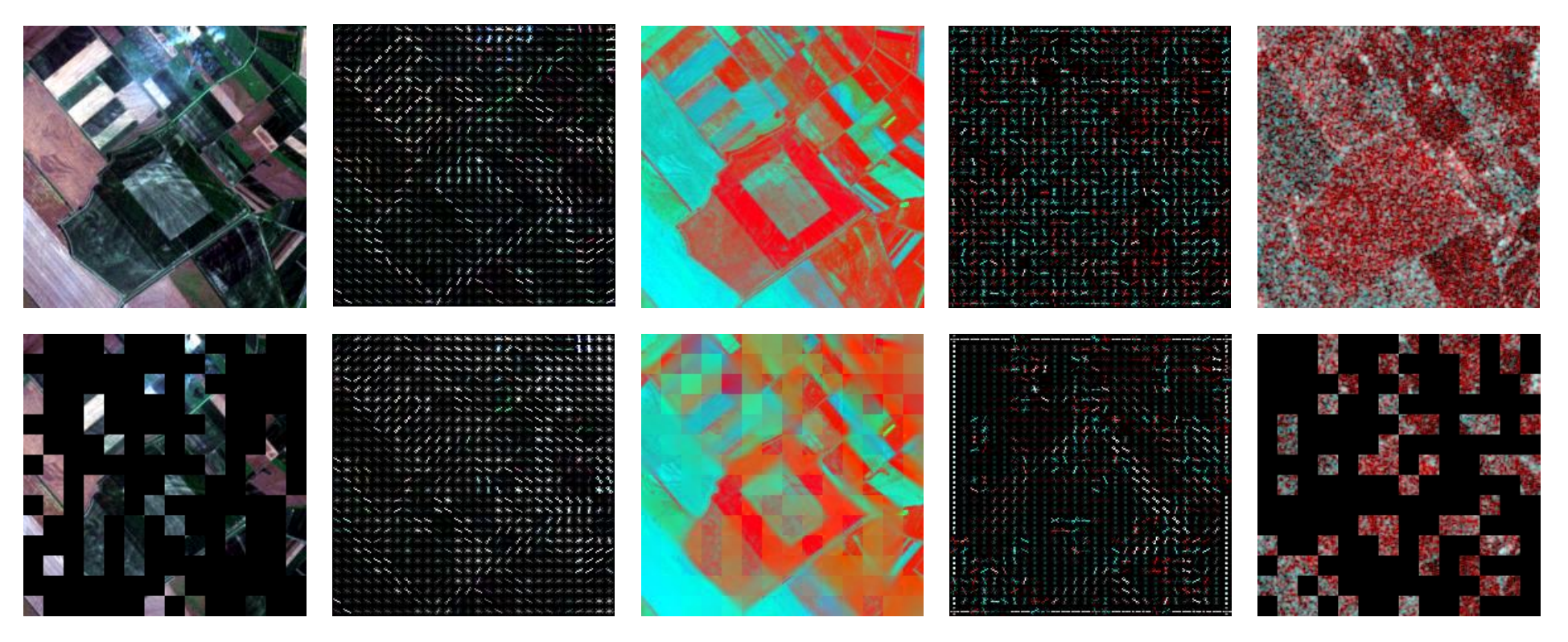}
    \includegraphics[width=1.0\linewidth]{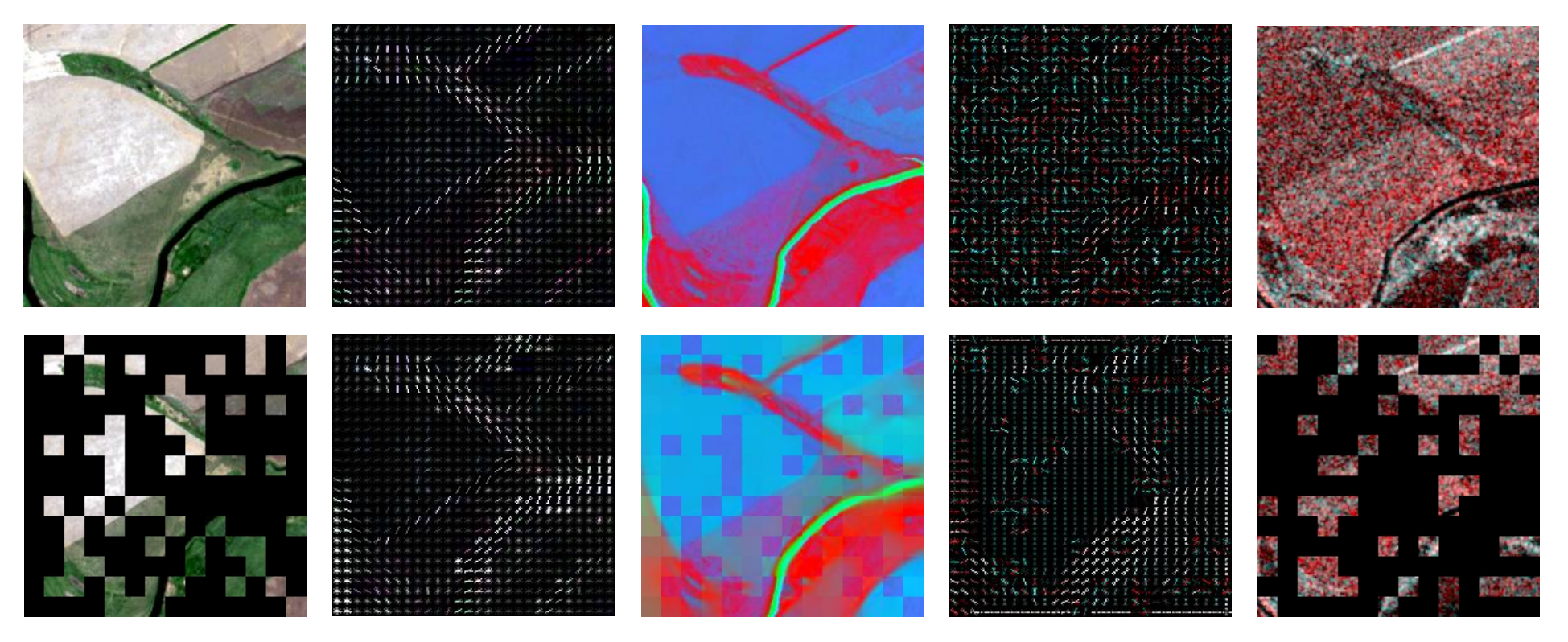}
    \includegraphics[width=1.0\linewidth]{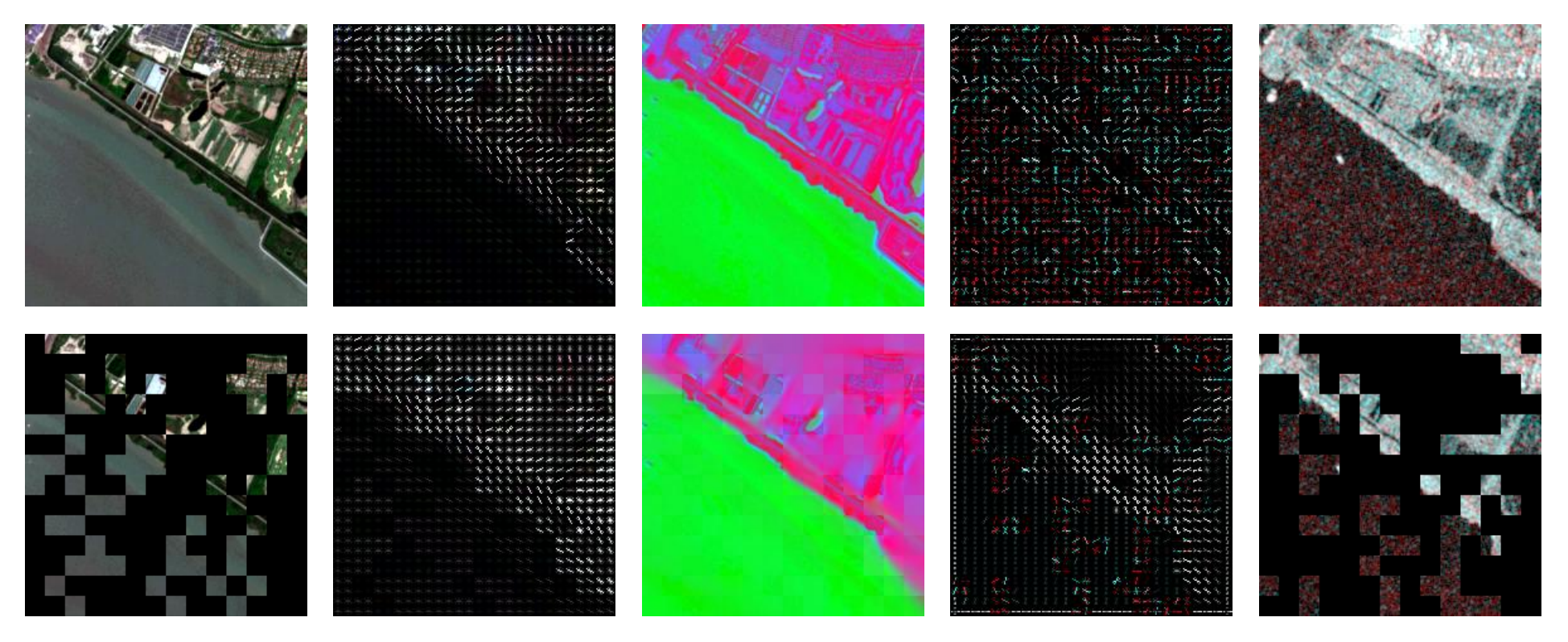}    
    
    \caption{Examples of FG-MAE reconstructed features. Every two rows represent one MS-SAR pair. From left to right, first row: MS image, MS HOG target, MS NDI target, SAR HOG target, SAR image; second row: MS image masked, MS HOG prediction, MS NDI reconstruction, SAR HOG prediction, SAR image masked.
    }
    \label{fig:features-more}
\end{figure*}

\vspace{-0.5em}
\begin{table}[h!]
\centering
\caption{A study of the features on BigEarthNet-10\% -- MS.}
\label{tab:feature-ms}
\begin{tabular}{lcc}
\toprule
                  & Linear probing & Fine tuning \\ \toprule
Rand. Init.       & 70.3                  & -           \\
Supervised        & -                     & 81.3        \\ \cdashline{1-3}
Raw image (MAE)         & 77.8                  & 84.8        \\
CannyEdge         & 77.9                  & 84.8        \\
HOG               & 77.9                  & 85.0        \\
Dense SIFT              & 77.8  
& 84.9            \\
NDI               & 77.3                  & 84.6        \\ \cdashline{1-3}
HOG\&NDI (ours)         & \textbf{78.1}         & \textbf{85.2}        \\ \bottomrule
\end{tabular}
\end{table}

Among the individual features, both CannyEdge and HOG show an advantage over NDI in linear probing. This is due to the fact that spatial feature descriptors capture better image-level semantics from e.g. shape information, while the spectral feature NDI does not consider pixel relationships. In addition, while HOG performs best among individual features, NDI provides a good complement that combining both pushes the performances further.

A similar but more interesting behavior is shown in Table \ref{tab:feature-sar} for SAR imagery. We can observe that both SIFT and HOG perform better than raw image (MAE), and HOG provides a remarkable boost. This can be attributed to the fact that MAE reconstructs every pixel and thus strongly disturbed by the speckle noise, while feature descriptors provide natural noise filtering. Furthermore, Dense SIFT performs worse than HOG. This is due to the coarse setting that we consider each pixel as one key point and thus have too many false positives. In fact, this inspires a promising research direction to better integrate scale-invariant features into MAE structure.

\begin{table}[h!]
\centering
\caption{A study of the features on BigEarthNet-10\% -- SAR.}
\label{tab:feature-sar}
\begin{tabular}{lcc}
\toprule
                  & Linear classification & Fine tuning   \\ \toprule
Rand. Init.       & 58.1                  & -             \\
Supervised        & -                     & 72.7          \\ \cdashline{1-3}
Raw image (MAE)         & 69.8                  & 74.9          \\
CannyEdge         & 69.9                   & 74.9           \\
Dense SIFT              & 69.8                  & 75.8
     \\
HOG (ours)             & \textbf{71.7}         & \textbf{78.0} \\ \bottomrule
\end{tabular}
\end{table}

Qualitative examples of feature reconstruction can be seen in Figure \ref{fig:features} and \ref{fig:features-more}. Despite masking out 70\% of the input patches, the reconstruction results remain impressive for multispectral images. For SAR images, the ground truth themselves are very noisy, but interestingly, the reconstructed features appear clearer than the ground truth. This observation suggests another exciting research direction for better low-level feature extraction algorithms \cite{dellinger2014sar} .

\subsection{FGMAE-MS}
We then benchmark the performance of the proposed FGMAE-MS (HOG+NDI) and FGMAE-SAR (HOG) on extensive downstream datasets. Table \ref{tab:BE-ms} shows the transfer results on the full set of the multi-label scene classification dataset BigEarthNet. The proposed FGMAE-MS outperforms MAE consistently on both linear probing and fine tuning, with improvements up to 0.9\%.

\vspace{-0.4em}
\begin{table}[h]
\centering
\caption{FGMAE-MS on BigEarthNet-100\%.}
\label{tab:BE-ms}
\begin{tabular}{lcccc}
\toprule
                       & \multicolumn{2}{c}{Linear classification} & \multicolumn{2}{c}{Fine tuning} \\
                       & mAP                 & F1                  & mAP            & F1             \\ \toprule
Rand. Init.            & 72.0                & 60.0                & -              & -              \\
Supervised             & -                   & -                   & 87.8           & 78.9           \\ \cdashline{1-5}
MAE                    & 78.0                & 68.0                & 88.6           & 79.9           \\
FG-MAE (ours)           & \textbf{78.5}       & \textbf{68.7}       & \textbf{89.3}  & \textbf{80.8}  \\ 
\bottomrule
\end{tabular}
\end{table}

Table \ref{tab:EU-ms} presents the transfer learning results on the single-label scene classification dataset EuroSAT. Similar to BigEarthNet, slight but consistent improvements can be observed in all scenarios.

\vspace{-0.4em}
\begin{table}[h]
\centering
\caption{FGMAE-MS on EuroSAT.}
\label{tab:EU-ms}
\begin{tabular}{lcccc}
\toprule
                       & \multicolumn{2}{c}{Linear classification}    & \multicolumn{2}{c}{Fine tuning}        \\
                       & OA            & AA            & OA            & AA            \\ \toprule
Rand. Init.            & 79.3          & 79.5          & -             & -             \\
Supervised             & -             & -             & 96.7          & 96.3          \\ \cdashline{1-5}
MAE                    & 94.2          & 94.0          & 98.5          & 98.2          \\
FG-MAE (ours) & \textbf{94.8} & \textbf{94.8} & \textbf{98.7} & \textbf{98.5} \\ \bottomrule
\end{tabular}
\end{table}

Finally, Table \ref{tab:dfc-ms} demonstrates the transfer learning results on the semantic segmentation dataset DFC2020, where FGMAE-MS outperforms MAE by noticeable margins on all metrics (e.g. 3.4\% increase in mIoU). This underscores the promising benefits of FG-MAE on dense prediction tasks.

\vspace{-0.4em}
\begin{table}[h!]
\centering
\caption{FGMAE-MS on DFC2020.}
\label{tab:dfc-ms}
\begin{tabular}{lccc}
\toprule
                       & OA            & mIoU          & AA          \\ \toprule
Supervised             & 63.3           & 46.2          & 59.2         \\ \cdashline{1-4}
MAE                    & 66.9           & 48.0          & 63.5         \\
FG-MAE (ours) & \textbf{69.6} & \textbf{51.4} & \textbf{66.4} \\ \bottomrule
\end{tabular}
\end{table}

\begin{figure*}[h]
    \centering
    
    \includegraphics[width=1.0\linewidth]{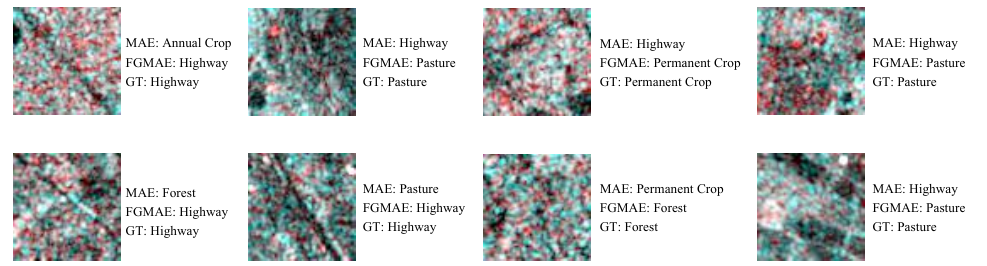}
    
    \caption{Examples of EuroSAT-SAR prediction results where FG-MAE gives the correct label while MAE doesn't. FG-MAE better captures semantics that are more distinguishable from the HOG features (e.g. a highway).}
    \label{fig:eu-sar-wrong}
\end{figure*}

\subsection{FGMAE-SAR}

\begin{table*}[]
\centering
\caption{Per-class benchmark results on EuroSAT-SAR. FG-MAE outperforms MAE by large margins on most of the classes.}
\label{tab:EU-sar-cls}
\scalebox{0.85}{
\begin{tabular}{lccccccccccc}
\toprule
             & Annual Crop & Forest & Herbaceous Vegetation & Highway & Industrial & Pasture & Permanent Crop & Residential           & River & Sea/Lake \\ \toprule
Supervised   & 76.4            & 77.7       & 66.4                      & 66.0        & 90.7           & 58.4     & 59.1               & 90.2     & 89.3      & 98.1       \\
MAE          & 79.6            & 79.1       & 70.2                      & 72.7        & 92.0           & 64.6     & 59.2               & 91.8                      & 92.4      & \textbf{98.8}        \\
FG-MAE (ours) & \textbf{84.1} (+3.5)            & \textbf{85.4} (+6.3)        & \textbf{78.1} (+7.9)                      & \textbf{82.4} (+9.7)        & \textbf{93.5} (+1.5)        & \textbf{75.7} (+11.1)         & \textbf{67.7} (+8.5)               & \textbf{94.3} (+2.5)                      & \textbf{94.2} (+1.8)      & \textbf{98.8}       \\ \hline
\end{tabular}
}
\end{table*}

Likewise, we benchmark the transfer learning results on SAR imagery of the aforementioned datasets. As can be seen from Table \ref{tab:BE-sar}, FGMAE-SAR demonstrates remarkable improvements compared to MAE on BigEarthNet. Especially when compared to the multispectral scenario (0.5\% to 0.9\% improvements), the benefit of FG-MAE is much more significant (up to 3.1\%). This again highlights the advantage of implicit noise filtering with HOG features.

\begin{table}[h]
\centering
\caption{FGMAE-SAR on BigEarthNet-100\%.}
\label{tab:BE-sar}
\begin{tabular}{lcccc}
\toprule
                       & \multicolumn{2}{c}{Linear classification} & \multicolumn{2}{c}{Fine tuning} \\
                       & mAP                 & F1                  & mAP            & F1             \\ \toprule
Rand. Init.            & 59.0                & 40.4                & -              & -              \\
Supervised             & -                   & -                   & 79.5           & 71.1           \\ \cdashline{1-5}
MAE                    & 70.4                & 59.1                & 81.3           & 72.8           \\
FG-MAE (ours)           & \textbf{72.3}       & \textbf{62.2}       & \textbf{82.7}  & \textbf{74.0}  \\ 
\bottomrule
\end{tabular}
\end{table}

Table \ref{tab:EU-sar} presents the results on our collected EuroSAT-SAR dataset. Similar to BigEarthNet results, substantial performance boosts can be observed with FGMAE-SAR. While FGMAE-MS gives 0.2\% to 0.8\% improvements compared to MAE, FGMAE-SAR provides up to 5.0\% improvement. Detailed per-class benchmarks are shown in Table \ref{tab:EU-sar-cls}, where FGMAE-SAR outperforms MAE by a large margin for most of the classes (e.g. as much as 11.1\% for the pasture class). Figure \ref{fig:eu-sar-wrong} presents some patch examples, which MAE misclassifies while FG-MAE predicts the correct label. We can observe from the figure that FG-MAE helps the model better capture the semantics that are easier to recognize with HOG features (e.g. a highway image).

\begin{table}[h]
\centering
\caption{FGMAE-SAR on EuroSAT.}
\label{tab:EU-sar}
\begin{tabular}{lcccc}
\toprule
                       & \multicolumn{2}{c}{Linear classification}    & \multicolumn{2}{c}{Fine tuning}        \\
                       & OA            & AA            & OA            & AA            \\ \toprule
Rand. Init.            & 61.9          & 61.3          & -             & -             \\
Supervised             & \textbf{-}    & \textbf{-}    & 78.4          & 77.7          \\ \cdashline{1-5}
MAE                    & 79.3          & 78.6          & 81.0          & 80.4          \\
FG-MAE (ours) & \textbf{80.7} & \textbf{79.9} & \textbf{85.9} & \textbf{85.4} \\ \bottomrule
\end{tabular}
\end{table}

\begin{figure*}[]
    \centering
    
    \includegraphics[width=1.0\linewidth]{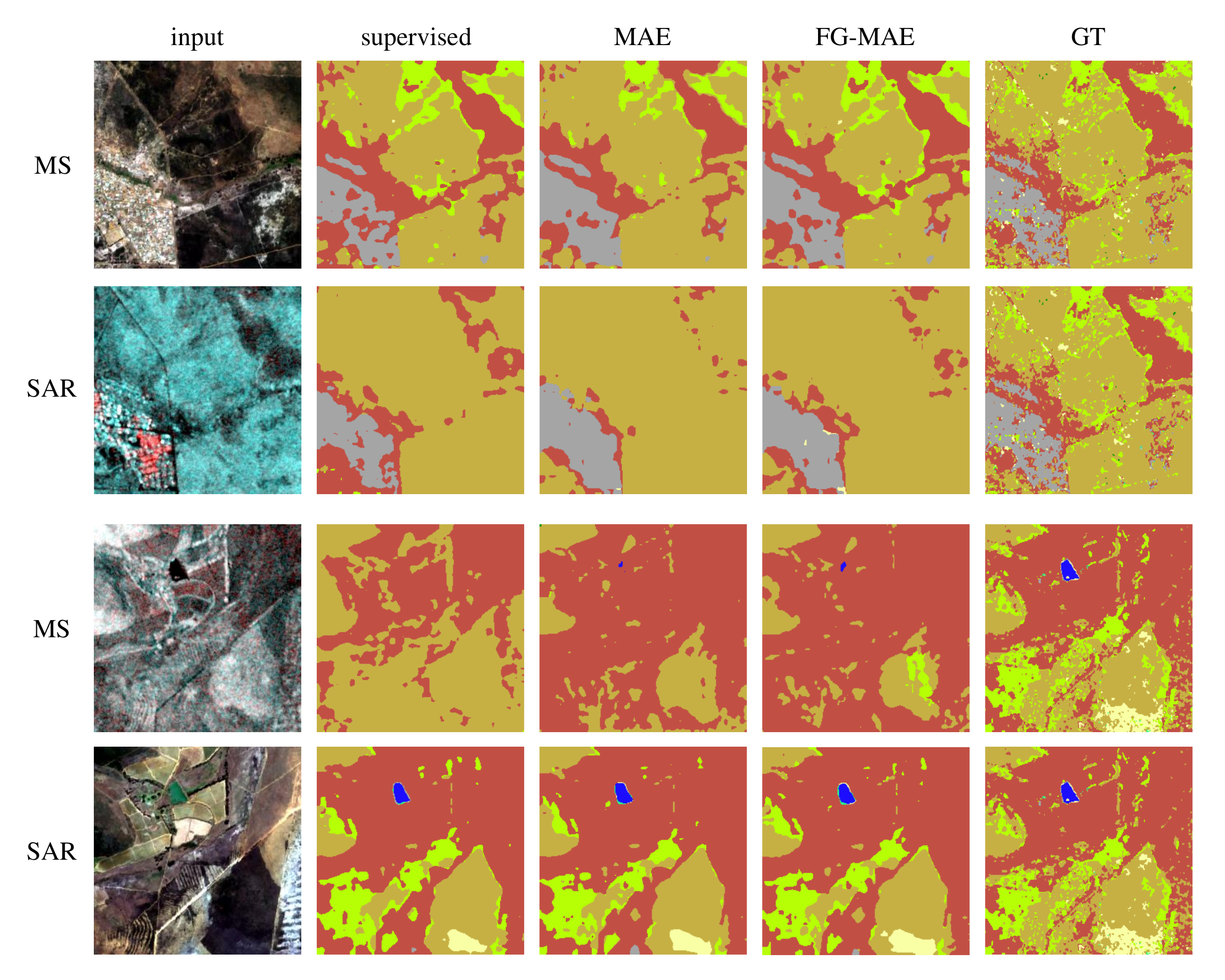}
    
    \caption{Examples of DFC2020 segmentation maps. Every two rows represent one MS-SAR pair. From left to right, first row: MS image, MS prediction supervised, MS prediction MAE, MS prediction FG-MAE, ground truth mask; second row: SAR image, SAR prediction supervised, SAR prediction MAE, SAR prediction FG-MAE, ground truth mask.}
    \label{fig:dfc2020}
\end{figure*}

Finally on DFC2020, consistent improvements compared to MAE can be seen from Table \ref{tab:dfc-sar}. Though the improvements compared to FGMAE-MS here are not as much as the previous two scene classification datasets, they are still noteworthy compared to supervised learning. This is also shown in Figure \ref{fig:dfc2020}, where the segmentation results of two example image pairs are presented. The limited benefits can be attributed to the characteristics of SAR imagery, where interpreting fine grained pixel details is very challenging.  

\begin{table}[h]
\centering
\caption{FGMAE-SAR on DFC2020.}
\label{tab:dfc-sar}
\begin{tabular}{lccc}
\toprule
                       & OA            & mIoU          & AA          \\ \toprule
Supervised             & 61.4          & 37.3          & 56.1          \\ \cdashline{1-4}
MAE                    & 62.1          & 38.9          & 56.9          \\
FG-MAE (ours) & \textbf{62.3} & \textbf{39.3} & \textbf{57.0} \\ \bottomrule
\end{tabular}
\end{table}

\subsection{Scaling ViTs}

\begin{figure}[]
  \centering
  \subfigure[FGMAE-MS]{\includegraphics[width=0.9\linewidth]{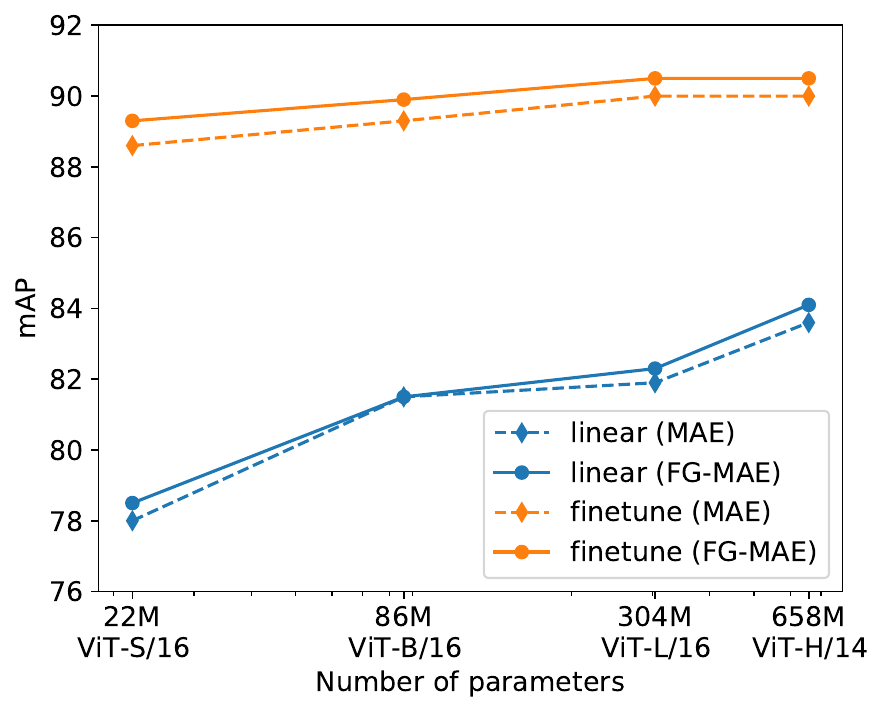}}
  \subfigure[FGMAE-SAR]{\includegraphics[width=0.9\linewidth]{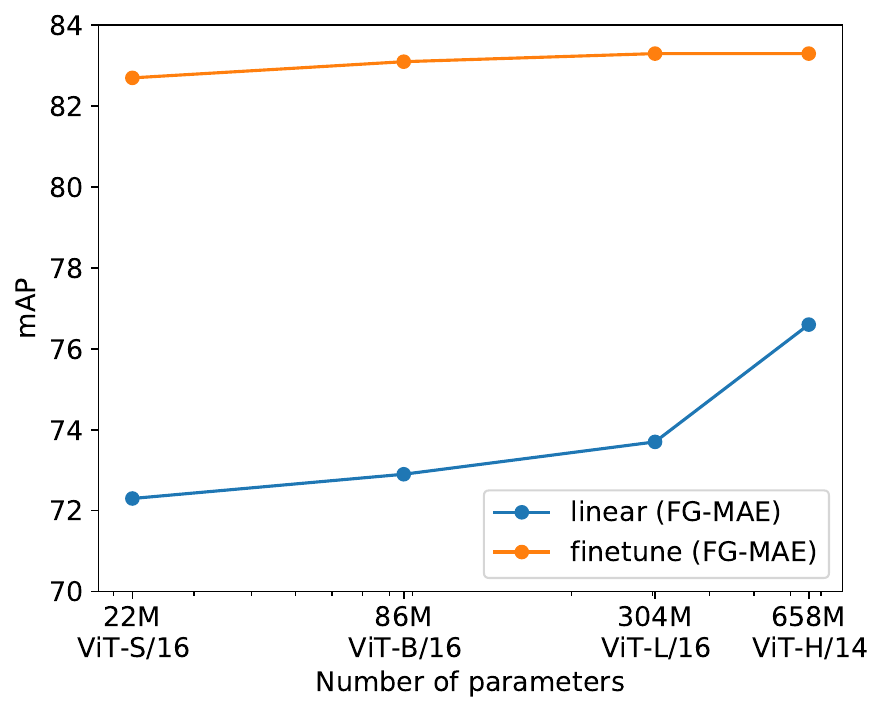}}
  \caption{Similar to MAE, FG-MAE scales well on BigEarthNet linear evaluation for both multispectral and SAR imagery.}
  \label{fig:scale}
\end{figure}

The efficiency of MAE is well-preserved in the proposed FG-MAE, thus we are able to scale-up the pretrained models to a series of ViTs with up to 658 million parameters: ViT-Small, ViT-Base, ViT-Large and ViT-Huge. We evaluate linear classification and fine tuning results on both multispectral and SAR imagery of the BigEarthNet-MM dataset. As is shown in Figure \ref{fig:scale}, scaling up ViTs provides consistent improvements for both modalities under linear classification protocol. This supports the potential benefits of even larger foundation models \cite{cha2023abillion}. However, we also observe significant overfitting phenomenon under fine tuning protocol, as reflected by the saturation trend in Figure \ref{fig:scale}. This indicates the need for further research on how to effectively fine-tune big foundation models.

\section{Conclusion}
\label{sec:conclusion}

In this study, we demonstrated that image features are comparable or superior reconstruction targets for masked image modelling based pretraining in remote sensing, particularly for SAR imagery. We proposed a novel variant of MAE, called feature guided masked autoencoder (FG-MAE), which modifies the reconstruction targets. For multispectral imagery, we combined HOG and NDI, while for SAR imagery, we used HOG alone. Experimental results on three downstream tasks verify the effectiveness of FG-MAE. In addition, we demonstrated the scalability of FG-MAE, and released a series of pretrained vision transformers with size up to 0.7B parameters for multispectral and SAR imagery.  

Though the proof of concept has been made clear, one limitation of this work is that we can not make the best use of scale-invariant features such as SIFT / SAR-SIFT out-of-the-box. However, we believe these features are of great value with proper and more sophisticated design. Scale-MAE \cite{reed2023scale}, for example, though not directly inspired by SIFT, shares a similar idea and provides promising insights.

Another limitation, as we have mentioned, is that both MAE and FG-MAE scale well to larger backbones in linear probing, but not in fine tuning. As we are entering the era of big EO foundation models, how to effectively transfer the foundation knowledge remains an important but not yet well-studied problem.

There are also two interesting thoughts that we believe deserve further investigation. First, we have shown a relatively poor performance in reconstructing raw SAR images because of the effect of speckle noise. However, what if we reconstruct the despeckled images instead? SAR-despeckling has been widely studied and there are many well-developed algorithms. If integrated into MAE pretraining, would it help the model prevent confusion due to noise? Second, the reconstructed SAR features sometimes seem to be clearer than the corresponding noisy ground truth. This may inspire a promising direction for low-level tasks, including the aforementioned SAR-despeckling.

\section*{Acknowledgement}

This work was funded by the Helmholtz Association through the Framework of Helmholtz AI, grant ID: \texttt{ZT-I-PF-5-01} -- \textit{Local Unit Munich Unit @Aeronautics, Space and Transport (MASTr)}. The compute was supported by the Helmholtz Association's Initiative and Networking Fund on the HAICORE@FZJ partition. The work of X. Zhu is supported by the German Federal Ministry of Education and Research (BMBF) in the framework of the international future AI lab "AI4EO -- Artificial Intelligence for Earth Observation: Reasoning, Uncertainties, Ethics and Beyond" (grant number: 01DD20001).


{
\printbibliography
}

\clearpage
\newpage

\begin{figure*}
    \centering
\begin{minipage}{\textwidth}
\centering
\large\textbf{Appendix: EuroSAT-SAR Dataset}
\end{minipage}
\end{figure*}


Below we provide extensive information about the collected EuroSAT-SAR dataset, which is a SAR version of EuroSAT. As a side contribution of this paper, we believe this simple, clean and well-balanced SAR dataset (which surprisingly rarely exists yet) is of great value to further machine-learning research and education on SAR imagery.

\vspace{-0.5em}
\subsection{\textbf{Data collection}}
\begin{figure}[h]
    \centering
    \includegraphics[width=0.75\linewidth]{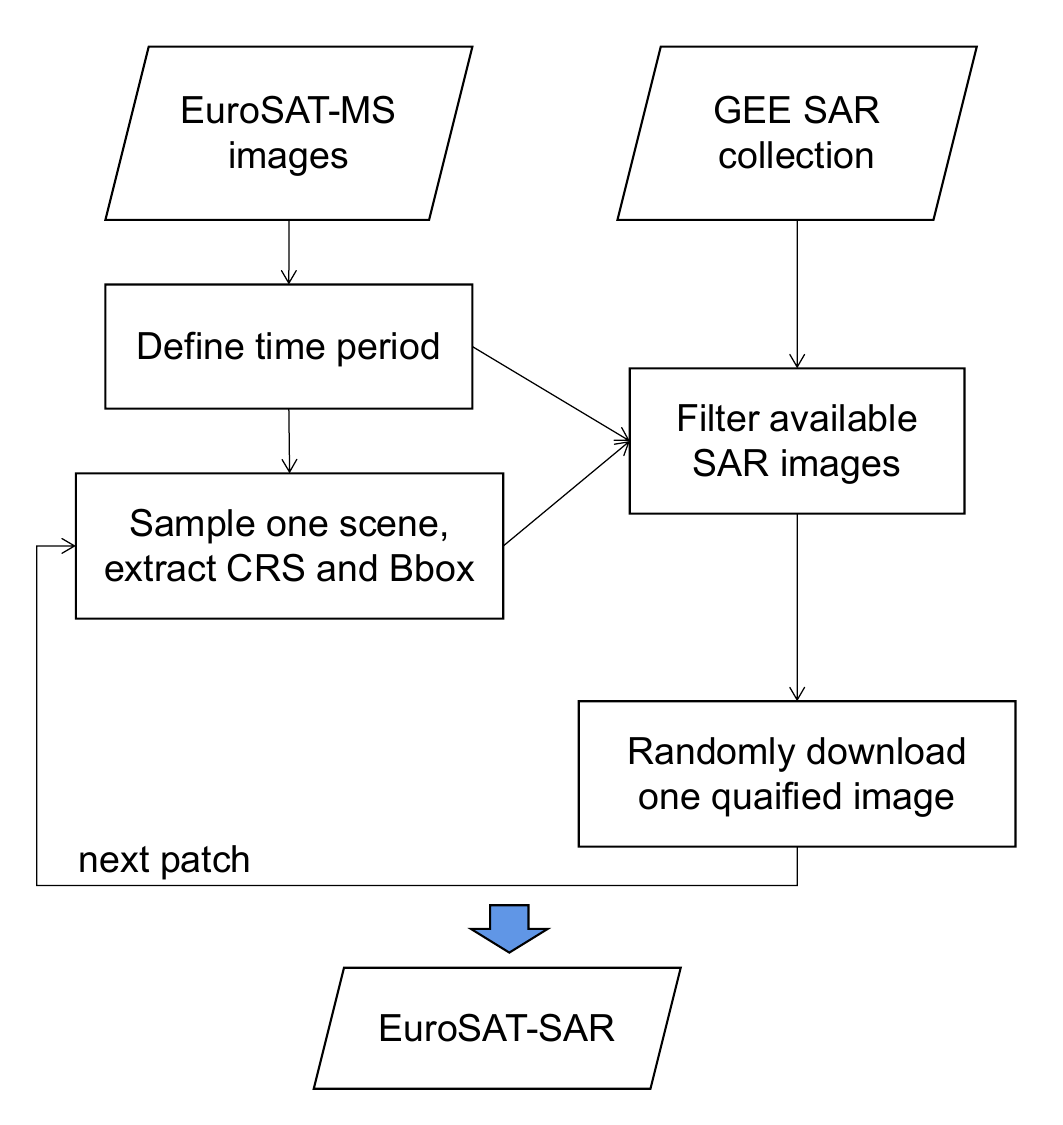}
    \caption{EuroSAT-SAR creation pipeline.}
    \label{fig:flowchart}
\end{figure}

To create EuroSAT-SAR, we match the published EuroSAT-MS (Sentinel-2 L1C) dataset with dual-pol Sentinel-1 GRD images from Google Earth Engine. Specifically, for each geotiff image in EuroSAT-MS, we extract the corresponding coordinate system and bounding box coordinates. We then build a geo-referenced rectangle region for the patch. Meanwhile, we build a temporal period between the years 2016 and 2017. Next, we filter available SAR images based on the region and period, and download a random qualified patch with bands VV and VH. Since no cloud filtering is needed for SAR imagery, the data collection is very fast (within one hour). In the end, we match the whole EuroSAT-MS dataset and download 27,000 SAR images, each assigned with the same class label as the corresponding MS image. Figure \ref{fig:flowchart} illustrates the data collection process.

\vspace{-0.5em}

\vspace{-1em}
\subsection{\textbf{Dataset characteristics}}

\begin{figure}[h!]
    \centering
    
    \subfigure[Annual Crop]{
    \includegraphics[width=0.36\linewidth]{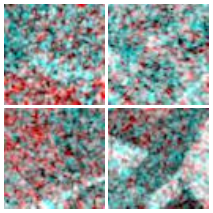}}
    \vspace{-0.5em}
    \subfigure[Forest]{
    \includegraphics[width=0.36\linewidth]{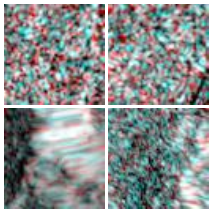}}  
    \vspace{-0.5em}
    \subfigure[Herbaccous Vegetation]{
    \includegraphics[width=0.36\linewidth]{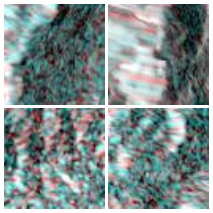}}    
    \subfigure[Highway]{
    \includegraphics[width=0.36\linewidth]{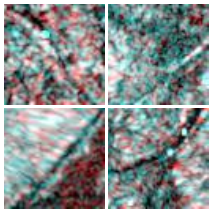}} 
    \vspace{-0.5em}
    \subfigure[Industrial]{
    \includegraphics[width=0.36\linewidth]{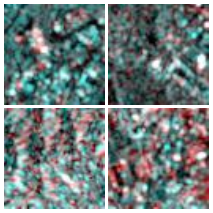}}    
    \subfigure[Pasture]{
    \includegraphics[width=0.36\linewidth]{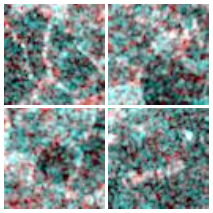}}
    \vspace{-0.5em}
    \subfigure[Permanent Crop]{
    \includegraphics[width=0.36\linewidth]{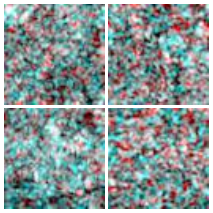}}    
    \subfigure[Residential]{
    \includegraphics[width=0.36\linewidth]{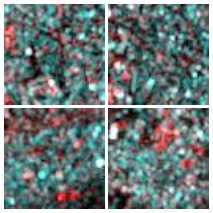}}
    \vspace{-0.5em}
    \subfigure[River]{
    \includegraphics[width=0.36\linewidth]{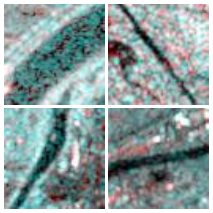}}    
    \subfigure[Sea/Lake]{
    \includegraphics[width=0.36\linewidth]{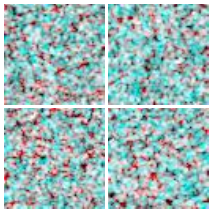}}    
    
    \caption{Sample image patches of all 10 classes covered in the collected EuroSAT-SAR dataset.}
    \label{fig:eu-sar-example}
\end{figure}

\vspace{-0.5em}
\begin{table*}[hb]
\centering
\caption{EuroSAT-SAR class distribution.}
\label{tab:eu-class}
\begin{tabular}{ccccccccccc}
\toprule
                 & Annual Crop & Forest & Herbaceous Vegetation & Highway & Industrial & Pasture & Permanent Crop & Residential & River & Sea/Lake \\ \hline
Number of images & 3000        & 3000   & 3000                  & 2500    & 2500       & 2000    & 2500           & 3000        & 2500  & 3000 \\   
\bottomrule
\end{tabular}
\end{table*}

EuroSAT-SAR dataset has 27,000 dual-pol Sentinel-1 GRD images with size 64$\times$64 and two channels VV and VH. There are 10 land cover land use classes, each containing 2000 to 3000 images. To complement the EuroSAT paper, Table \ref{tab:eu-class} presents the detailed class distribution. Also, sample images are shown in Figure \ref{fig:eu-sar-example}.


\clearpage
\newpage

\begin{figure*}
    \centering
\begin{minipage}{\textwidth}
\centering
\large\textbf{Datasheets for EuroSAT-SAR}
\end{minipage}
\end{figure*}

\setcounter{subsection}{0}
Here we answer the questions outlined in the datasheets for datasets paper by Gebru et al. \cite{gebru2021datasheets}.

\vspace{-3em}
\subsection{Motivation}
\vspace{-0.5em}

\textbf{For what purpose was the dataset created?} The dataset was created as a SAR version of the popular EuroSAT dataset to evaluate SAR foundation models.

\textbf{Who created the dataset (e.g., which team, research group) and on behalf of which entity (e.g.,
company, institution, organization)?} The dataset was created by the lab "Data Science in Earth Observation" at Technical University of Munich and German Aerospace Center.

\textbf{Who funded the creation of the dataset?} The creation of the dataset was funded by the Helmholtz Association through the Framework of Helmholtz AI.

\vspace{-2.5em}
\subsection{Composition}
\vspace{-0.5em}

\textbf{What do the instances that comprise the dataset represent (e.g., documents, photos, people, countries)?} This dataset contains satellite images.

\textbf{How many instances are there in total (of each type, if appropriate)?} The dataset contains 27,000 dual-pol SAR images with size 64$\times$64.

\textbf{Does the dataset contain all possible instances or is it a sample (not necessarily random) of instances from a larger set?} The dataset is a sample of all Sentinel-1 satellite images to match the EuroSAT dataset.

\textbf{What data does each instance consist of?} A Sentinel-1 GRD image.

\textbf{Is there a label or target associated with each instance?} Yes, the images are stored in different folders and labels are indicated by the folder names.

\textbf{Is any information missing from individual instances?} No.

\textbf{Are relationships between individual instances made explicit (e.g., users’ movie ratings, social network links)?} Not applicable, geographic location can be extracted if needed.

\textbf{Are there recommended data splits (e.g., training, development/validation, testing)?} No. Following EuroSAT dataset which doesn't have an official split, we provide the full dataset in a whole as well. We use our random splits in the benchmarks.

\textbf{Are there any errors, sources of noise, or redundancies in the dataset?} Yes, since we don't have the exact acquisition dates of EuroSAT images, we match them with SAR images in a rough time period assuming no change happened. Though the data looks good with some manual check, we didn't check all the images.

\textbf{Is the dataset self-contained, or does it link to or otherwise rely on external resources (e.g.,
websites, tweets, other datasets)?} The dataset is self-contained.

\textbf{Does the dataset contain data that might be considered confidential (e.g., data that is protected by legal privilege or by doctor-patient confidentiality, data that includes the content of individuals’ non-public communications)?} No.

\textbf{Does the dataset contain data that, if viewed directly, might be offensive, insulting, threatening, or might otherwise cause anxiety?} No.

\textbf{Does the dataset identify any subpopulations (e.g., by age, gender)?} No.

\textbf{Is it possible to identify individuals (i.e., one or more natural persons), either directly or indirectly (i.e., in combination with other data) from the dataset?} No.

\textbf{Does the dataset contain data that might be considered sensitive in any way (e.g., data that reveals race or ethnic origins, sexual orientations, religious beliefs, political opinions or union memberships, or
locations; financial or health data; biometric or genetic data; forms of government identification, such as social security numbers; criminal history)?} No.

\subsection{Collection process}

\textbf{How was the data associated with each instance acquired?} The data was collected from the publicly available Sentinel-1/2 database. 

\textbf{What mechanisms or procedures were used to collect the data (e.g., hardware apparatus or
sensor, manual human curation, software program, software API)?} Google Earth Engine with Python was used to collect the data.

\textbf{If the dataset is a sample from a larger set, what was the sampling strategy (e.g., deterministic, probabilistic with specific sampling probabilities)?} We sample Sentinel-1 images by matching the geocoordinates and rough acquisition time of the published EuroSAT dataset.

\textbf{Who was involved in the data collection process (e.g., students, crowdworkers, contractors) and how were they compensated (e.g., how much were crowdworkers paid)?} The data was automatically collected and verified by the authors.

\textbf{Over what timeframe was the data collected?} The data was collected by the authors between February and March 2022. The images within the dataset were captured in the year 2016/2017.

\textbf{Were any ethical review processes conducted (e.g., by an institutional review board)?} No.

\textbf{Did you collect the data from the individuals in question directly, or obtain it via third parties or other sources (e.g., websites)?} The data was collected from open sources.

\textbf{Were the individuals in question notified about the data collection?} N/A.

\textbf{Did the individuals in question consent to the collection and use of their data?} N/A.

\textbf{If consent was obtained, were the consenting individuals provided with a mechanism to revoke their consent in the future or for certain uses?} N/A.

\textbf{Has an analysis of the potential impact of the dataset and its use on data subjects (e.g., a data protection impact analysis) been conducted?} N/A.

\subsection{Preprocessing/cleaning/labeling} 

\textbf{Was any preprocessing/cleaning/labeling of the data done (e.g., discretization or bucketing, tokenization, part-of-speech tagging, SIFT feature extraction, removal of instances, processing of missing values)?} The data was pre-processed by GEE internally during the collection/downloading process. No further pre-processing was done.

\textbf{Was the “raw” data saved in addition to the preprocessed/cleaned/labeled data (e.g., to support unanticipated future uses)?} No, not necessary.

\textbf{Is the software used to preprocess/clean/label the instances available?} Yes, we use Google Earth Engine with Python which is freely available.

\subsection{Uses}

\textbf{Has the dataset been used for any tasks already?} In this paper we use the dataset as a downstream task to evaluate our proposed pretraining algorithms. 

\textbf{Is there a repository that links to any or all papers or systems that use the dataset?} Yes we will organize and maintain all related information at \url{https://huggingface.co/datasets/wangyi111/EuroSAT-SAR}.

\textbf{What (other) tasks could the dataset be used for?} The dataset can be used as a simple, clean SAR scene classification dataset for the remote sensing community, matching the popular multispectral EuroSAT dataset.

\textbf{Is there anything about the composition of the dataset or the way it was collected and preprocessed/cleaned/labeled that might impact future uses?} We do not unify the orbiting (ascending/descending) of Sentinel-1 data, which should be taken into consideration for SAR related applications.

\textbf{Are there tasks for which the dataset should not be used?} The authors are not aware of any specific task that should be avoided.

\subsection{Distribution}

\textbf{Will the dataset be distributed to third parties outside of the entity (e.g., company, institution, organization) on behalf of which the dataset was created?} Yes, the dataset is publicly available.

\textbf{How will the dataset will be distributed (e.g., tarball on website, API, GitHub)?} The dataset will be distributed as tarball. Access to the dataset can be found at \url{https://huggingface.co/datasets/wangyi111/EuroSAT-SAR}.

\textbf{When will the dataset be distributed?} Starting from July 2023.

\textbf{Will the dataset be distributed under a copyright or other intellectual property (IP) license, and/or under applicable terms of use (ToU)? } MIT license.

\textbf{Have any third parties imposed IP-based or other restrictions on the data associated with the instances?} No.

\textbf{Do any export controls or other regulatory restrictions apply to the dataset or to individual
instances?} No.

\subsection{Maintenance} 

\textbf{Who is supporting/hosting/maintaining the dataset?} The dataset is supported and maintained by the authors.

\textbf{How can the owner/curator/manager of the dataset be contacted (e.g., email address)?} The manager of the dataset can be reached at the email addresses: yi4.wang@tum.de or yi.wang@dlr.de.

\textbf{Is there an erratum?} If errors are found an erratum will be added.

\textbf{Will the dataset be updated (e.g., to correct labeling errors, add new instances, delete instances)?} Any updates will be posted and the dataset will be versioned.

\textbf{If the dataset relates to people, are there applicable limits on the retention of the data associated with the instances (e.g., were individuals in question told that their data would be retained for a fixed period of time and then deleted)?} N/A.

\textbf{Will older versions of the dataset continue to be supported/hosted/maintained?} Depending on the updates (if there are), we will either continue hosting the older versions or make a clear update log that older versions can be generated from the newest version.

\textbf{If others want to extend/augment/build on/contribute to the dataset, is there a mechanism for them to do so?} Yes, please feel free to reach out to us.

\subsection{Author statement of responsibility}

The authors confirm all responsibility in case of violation of rights and confirm the licence associated with the dataset.

\end{document}